\documentclass[10pt,twocolumn,letterpaper]{article}

\usepackage{cvpr}
\usepackage{times}
\usepackage{epsfig}
\usepackage{graphicx}
\usepackage{amsmath}
\usepackage{amssymb,comment}

\usepackage{subcaption}

\usepackage[pagebackref=true,breaklinks=true,letterpaper=true,colorlinks,bookmarks=false]{hyperref}

\cvprfinalcopy

\ifcvprfinal\fi
\begin{document}

\title{Harmonic Networks: Integrating Spectral Information into CNNs}

\author{Matej Ulicny
\qquad
Vladimir A. Krylov
\qquad
Rozenn Dahyot\\
ADAPT Centre, School of Computer Science and Statistics, Trinity College Dublin, Dublin, Ireland\\
{\tt\small \{ulinm,vladimir.krylov,rozenn.dahyot\}@tcd.ie}
}

\maketitle

\begin{abstract}
   Convolutional neural networks (CNNs) learn filters in order to capture local correlation patterns in feature space. In contrast, in this paper we propose harmonic blocks that produce features by learning optimal combinations of spectral filters defined by the Discrete Cosine Transform. The harmonic blocks are used to replace conventional convolutional layers to construct partial or fully harmonic CNNs. We extensively validate our approach and show that the introduction of harmonic blocks into state-of-the-art CNN baseline architectures results in comparable or better performance in classification tasks on small NORB, CIFAR10 and CIFAR100 datasets. 
\end{abstract}

\section{Introduction}

CNNs have been designed to take advantage of implicit characteristics of natural images, concretely correlation in local neighborhood and feature equivariance.
The wide application of features obtained by convolving images with explicitly defined local filters highlights the shift towards local learning from the extraction of global information. 

Wavelets --- one of the popular feature extractors, evaluate correlation of the input signal with localized waveforms. The signal is decomposed into its frequency spectrum to facilitate further analysis. The scattering network proposed in~\cite{Bruna13} uses multiple layers of wavelet filters to model geometrical visual information. It has been shown that the scattering network built on complex-valued Morlet wavelets could achieve state of the art results in handwritten digit recognition and texture classification. The scattering network with its filters designed to extract translation and rotation invariant representations was shown to achieve comparable classification accuracy to unsupervised CNNs~\cite{Oyallon15}. 

The Discrete Cosine Transform (DCT) is used in JPEG encoding to transform image blocks into their corresponding spectral representations in order to capture the most information with a small number of coefficients. Motivated by frequency separation and energy compaction properties of DCT, we propose a network that learns features by combining responses of window-based DCT with a small receptive field. The key difference between the scattering network and our method is that the former creates a new path for each wavelet filter used at every layer, which consequently increases the number of paths exponentially with the increase of the network's depth. The features along each path are computed independently and are considered jointly only at the classifier level. Our method, referred to as {\it harmonic network}, learns how to combine spectral coefficients at every layer to produce fixed size representation defined as a weighted sum of responses to DCT filters. 

The paper is organized as follows.
We review first the related works in Sec.~\ref{sec:soa}, and define the DCT used to construct harmonic networks in Sec.~\ref{sec:dct}. Our approach is presented in Sec.~\ref{sec:method} and assessed experimentally on image classification task on several popular image datasets in Sec.~\ref{sec:experiments}.

\section{State-of-the-art}
\label{sec:soa}

\subsection{DCT \& CNNs }

Several works consider combining spectral information with CNNs.
CNNs trained on DCT coefficients are frequently used in forensics, specially for detection of multiply compressed images. A common practice in several works~\cite{Wang16,Amerini17,Barni17} is to classify histograms of pre-selected DCT coefficients by 1-dimensional convolutional network. In another work~\cite{Li17} a multi-branch 2-dimensional CNN is trained on feature maps spanned by the first 20 AC coefficients (corresponding to non-zero frequencies in DCT) extracted from JPEG images.

A number of studies have investigated the use of spectral image representations for object recognition. DCT on small resolution images coupled with coefficient truncation was used to speed up training of shallow fully connected neural networks~\cite{Fu16} and fully connected sparse autoencoders~\cite{Zou14}. DCT features from the entire image were used to train Radial Basis Function Network for face recognition~\cite{Er05}. A significant convergence speedup and case-specific accuracy improvement have been achieved by applying DCT transform to early stage learned feature maps in shallow CNNs~\cite{Ghosh16} whereas the later stage convolutional filters were operating on a sparse feature spectral representation. Compressed images generated by convolutional autoencoder have been shown to reduce computational complexity when used for large scale image classification~\cite{Torfason18}. In~\cite{Ulicny17,Gueguen18} it was demonstrated how precomputed or JPEG-extracted DCT coefficients can be efficiently used to train classification CNNs. 

\subsection{Wavelets \& CNNs}

The scattering network~\cite{Oyallon17,Singh17} based on 2 layers of rotation and scale invariant wavelet transforms was used as an initial transform of the image data prior to CNN training, which led to comparable classification accuracy with deeper models having similar number of parameters. Recent studies show that even the first-order scattering coefficients are sufficient to reduce the input representation and preserve discriminative information for training CNNs for various tasks~\cite{Oyallon18}. \textit{Williams et al.}~\cite{Williams16} have advocated image preprocessing with wavelet transform, but used different CNN for each frequency subband. Wavelet filters were also used as a preprocessing method prior to NN-based classifier~\cite{Said16}, and to enhance edge information in images prior to classification~\cite{Silva18}. 

Other works have used wavelets in CNN computational graphs. Second order coefficients from Fast Wavelet Transform were used in~\cite{Williams18} to design wavelet pooling operator. Similar approach was taken by \textit{Ripperl et al.} who designed spectral pooling~\cite{Rippel15} based on Fast Fourier Transform of the features and high-frequency coefficient truncation. They also proposed to parametrize filters in Fourier domain to decrease their redundancy and speed up the convergence. In both works, the pooled features were recovered with Inverse Fast Wavelet or Discrete Transform respectively, thus the CNN still operates in spatial domain. To address texture classification, \textit{Fujieda et al.}~\cite{Fujieda17} proposed a Wavelet Convolutional Network that is trained on responses of Haar wavelets and concatenates higher order coefficient maps along with features of the same dimensionality learned from lower-order coefficients. Similar approach is taken by \textit{Lu et al.}~\cite{Lu18} that learns from both spatial and spectral information that is decomposed from first layer features. The higher-order coefficients are also concatenated along with the lower dimensional feature maps. However, contrary to our method, Wavelet CNNs decompose only the input features and not features learned at intermediate stages. Moreover, the maximum number of decompositions performed was limited to the number of spatial resolutions of CNN feature maps. \textit{Worrall et al.} incorporated complex circular harmonics into CNNs to learn rotation equivariant representations~\cite{Worrall17}. Similarly to our harmonic block, the structured receptive field block~\cite{Jacobsen16} learns new filters by combining fixed filter set, but uses considerably larger set of Gaussian derivative basis.

The Fisher vectors (FV) is one of the standard pre-CNN feature extraction techniques for image recognition and retrieval which is often used in combination with linear classifiers such as linear SVM. Representations produced by the Deep Fisher network and a CNN were shown to be complementary~\cite{Simonyan13}, and a simple multiplication of their posteriors was shown to significantly improve performance of each method. Combining FV that encodes SIFT local image descriptors with nonlinear fully-connected network was demonstrated to surpass performance of classical FV-based pipelines~\cite{Perronnin15}.

\subsection{Compressing DNNs}

Numerous works have focused on compressing the size of neural networks and decreasing the inference and training time. Speedup and memory saving for inference can be achieved by approximating the trained full-rank CNN filters by separable rank-1 filters~\cite{Jaderberg14}. Alternatively, replacing expensive convolution operation in spatial domain by multiplication of transformed weights and filters in Fourier domain has been shown to speed up training even for small filters~\cite{Mathieu14}. Hashed Network (HashNet) architecture~\cite{Chen15} is designed by exploiting redundancy of the network's weights by randomly grouping weights into hash buckets, in which all the weights share the same parameter value. Assuming the learned filters are typically smooth, an extension called Frequency-Sensitive Hashed Network (FreshNet)~\cite{Chen16} groups weights based on their DCT representation. 

A different idea to compress a CNN was employed by \textit{Wang et al.}~\cite{Wang16b} where model weights after DCT transform are clustered and represented via shared cluster centers and particular weight residuals. Weights in this form were quantized and transformed via Huffman coding for storage purposes. Convolution was performed in the frequency domain to reduce the computational complexity. Slightly different pipeline~\cite{Han16} compresses the trained network by pruning small parameters, then clustering and quantizing all the weights. Pruned and quantized model weights were finetuned and then transformed by Huffman coding. It has been shown~\cite{Kim15} that a model complexity can be adjusted during the training time: increased via introduction of new filters by rotating and applying noise to existing ones, and reduced by clustering to selectively decrease their redundancy. A different approach~\cite{Denil13} for model compression preserves only a subset of model's parameters and predicts those that are missing.

\section{Discrete Cosine Transform} \label{sec:dct}

DCT is an orthogonal transformation method that decomposes an image to its spatial frequency spectrum. In continuous form, an image is projected to a sum of sinusoids with different frequencies. The contribution of each sinusoid towards the whole signal is determined by its coefficient calculated during the transformation.

It is a separable transform and due to its energy compaction properties it is commonly used for image and video compression in widely used JPEG and MPEG formats. DCT is one of the transformation methods with the best energy compaction properties on natural images~\cite{Yaroslavsky14}. Karhunen-Lo{\`e}ve transform is considered to be optimal in signal decorrelation, however it transforms signal via unique basis functions that are not separable and need to be estimated for every image. 

The literature provides different definitions of DCT. The most common DCT-II is computed on a 2-dimensional grid $X$ of size $A \times B$ representing the image patch with 1 pixel discretization step as
\begin{equation} \label{eq:dct}
\begin{split}
Y_{u,v} = & \sum_{x=0}^{A-1} \sum_{y=0}^{B-1}  \sqrt{\frac{\alpha_u}{A}} \sqrt{\frac{\alpha_v}{B}} X_{x,y} \times \\ & \cos{\left[\frac{\pi}{A} \left(x+\frac{1}{2}\right)u\right]} \cos{\left[\frac{\pi}{B}\left(y+\frac{1}{2} \right)v\right]}
\end{split}
\end{equation}
which results in coefficient $Y_{u,v}$ representing overlap of input with sinusoids at frequency $u$ and $v$ in particular directions. Basis functions are often normalized by scaling factors $\alpha_0=1$ and $\alpha_u=2, u>0$ to ensure orthogonality.

\section{Harmonic Networks}\label{sec:method}

A convolutional layer extracts correlation of input patterns with learned filters with local receptive fields. The idea of convolutions applied to images stems from the observation that pixels in local neighborhoods of natural images tend to be strongly correlated. In many image analysis applications, transformation methods are used to decorrelate signals forming an image. In contrast with spatial convolution with learned kernels, this study proposes feature learning by weighted combinations of predefined filter responses. These predefined filters extract harmonics from lower-level features in a local region.

We propose a network that consists of several {\it harmonic blocks} and optionally layers performing learned spatial convolution or fully-connected layers. A harmonic block processes data in 2 stages. In the first stage, the input features undergo harmonic decomposition by a transformation method. In this study we use window-based DCT, but can be replaced by other transformations, e.g. wavelets. In the second stage, transformed signals are combined by learned weights.

\begin{figure*}[t]
\begin{center}
\begin{center}
   \includegraphics[width=0.25\linewidth]{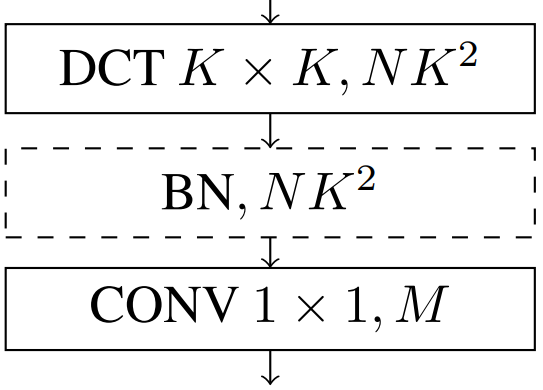}
   \includegraphics[width=0.7\linewidth]{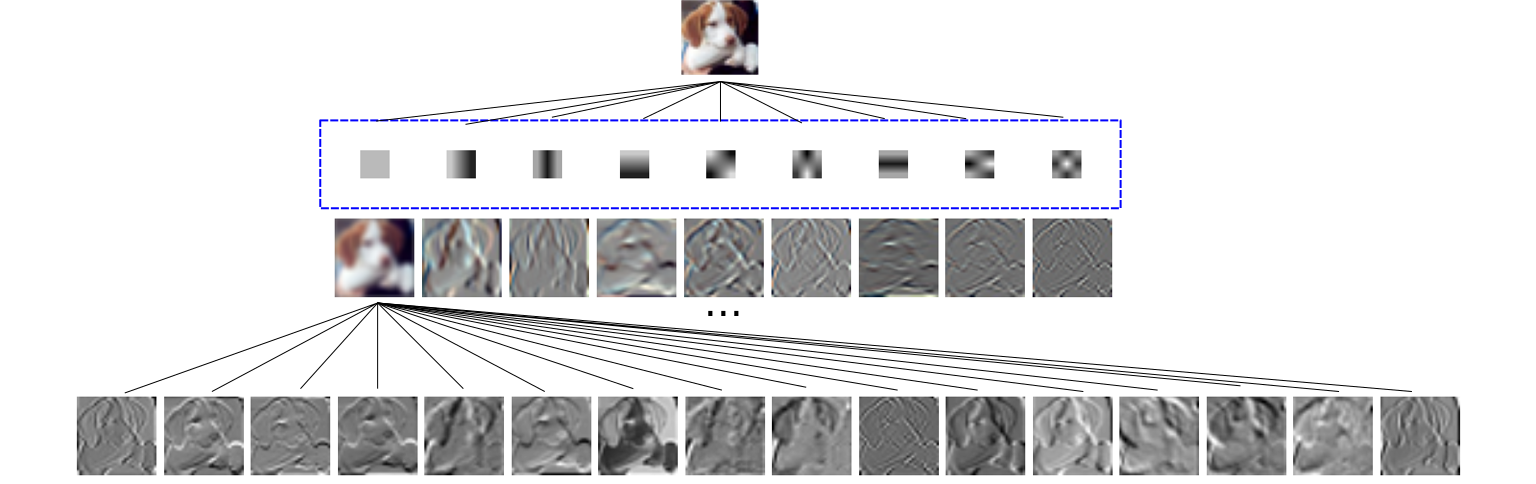}
\end{center}
\end{center}
\vspace{-1.25\baselineskip}
   \caption{Left: Design of a harmonic block that performs channel-wise windowed DCT transform and then recombines responses with 1x1 convolution. The boxes show the operations, sizes of the filters (if applicable) and the corresponding numbers of output channels given a set filter size $K$, number of input channels $N$ and output channels $M$. Batch normalization (BN) block is optional. Right: Visualization of the harmonic block applied to an input layer.}
\label{fig:harmlayer}
\end{figure*}

\subsection{Harmonic Block Design}\label{sec:method.design}

The harmonic block receives a set of 2-dimensional features $X$ as input. Note that the method is not restricted to this dimensionality, extension to data with different number of dimensions such as 1-dimensional sequences or 3-dimensional volumetric or spatio-temporal inputs is straightforward.
The block employs a transformation method to extract pixel neighborhood information from feature maps in terms of spatial frequency spectrum.

Spectral decomposition of input features into block-DCT representation is implemented as a convolution with DCT basis functions. A 2D kernel with size $K \times K$ is constructed for each basis function, comprising a filter bank of depth $K^2$, which is separately applied to each feature in the input. Convolution with the filter bank isolates coefficients of DCT basis functions to their exclusive feature maps, creating a new feature map per each channel and each frequency considered. The number of operations required to calculate this representation can be minimized by decomposing 2D DCT filter into two rank-1 filters and applying them as separable convolution to rows and columns sequentially. Despite the operation being computationally cheaper compared to dense convolutions, the spectral decomposition upsamples the number of intermediate features by $K^2$ factor, thus notably increasing the corresponding memory requirements.

The spectral representation of features is similar to a block-DCT representation of a JPEG image with a few notable differences. Blocks of an image or a feature transformed by DCT bases can overlap and the spectrum coefficients are mapped to the depth dimension, unlike mixing spatial and frequency information in JPEG. The depth dimension represents a set of feature maps composed of DCT coefficients, each corresponding to a particular DCT basis function. DCT feature maps of different input features are simply concatenated to build the spectral representation of the whole feature set. Each feature map $h^l$ at depth $l$ is computed as a weighted linear combination of DCT coefficients across all input channels $N$:
\begin{equation} \label{eq:feature}
  h^l = \sum_{n=0}^{N-1}{\sum_{u=0}^{K-1}{\sum_{v=0}^{K-1}{w^{l}_{n,u,v}\psi_{u,v}* *\, h^{l-1}_n}}}
\end{equation}
where $\psi_{u,v}$ is a $u,v$ frequency selective DCT filter of size $K \times K$, $**$ the 2-dimensional convolution operator and $w^{l}_{n,u,v}$ is learned weight for $u,v$ frequency of the $n$-th feature. The linear combination of spectral coefficients is implemented via a convolution with $1 \times 1$ filter that scales and sums the features, see Fig.~\ref{fig:harmlayer}.

Since the DCT is a linear transformation, backward pass through the transform layer is performed similarly to a backward pass through a convolution layer.

Harmonic block can be considered a special case of depth-separable convolution with predefined spatial filters. The fundamental difference from standard convolutional network is that the optimization algorithm is not searching for filters that extract spatial correlation, rather learns the relative importance of defined feature extractors (DCT filter responses) at multiple layers.

\subsection{Impact of Normalization} \label{sec:method.normal}

One of the key properties of the DCT transform is its energy compaction in low-frequency spectrum. It is expected that low frequency selective filters will have higher responses than the ones focusing on high frequencies. This effect can cause the model to focus on the responses of low-frequency filters and ignore output of some high-frequency filters. The loss of high frequency information throughout the network can be efficiently handled by normalizing spectrum of the input channels. The batch normalization is used to adjusts per frequency mean and variance, prior to the learned linear combination. The spectrum normalization transforms Eq.~\ref{eq:feature} into: 
\begin{equation} \label{eq:normalized_feature}
  h^l = \sum_{n=1}^{N-1}{\sum_{u=0}^{K-1}{\sum_{v=0}^{K-1}{w^l_{n,u,v}\frac{\psi_{u,v}** h^{l-1}_n - \mu^l_{n,u,v}}{\sigma^l_{n,u,v}}}}},
\end{equation}
with parameters $\mu^l_{n,u,v}$ and $\sigma^l_{n,u,v}$ estimated over the input batch rendering them dependant on the training batch-size.

\subsection{Harmonic Network Compression} \label{sec:method.subsample}

In standard JPEG encoding, block-DCT casts non-overlapping image blocks into the same number of coefficients as there were pixels in the original representation. The block size determines the receptive field for feature calculation. Block overlap prevents coarse spatial subsampling of features at the cost of increasing the volume of information produced by the transform. This trade-off is modulated by kernel size and stride while performing DCT transform via convolution. This form of subsampling is alternative to max-pooling, which has lower computational cost but discards substantial amounts of information. Response to the zero frequency DCT filter, the so called DC component, is equivalent to scaled average pooling with the same window size and stride as the DCT transform.

\begin{figure}[t]
\begin{center}
   \includegraphics[width=0.6\linewidth]{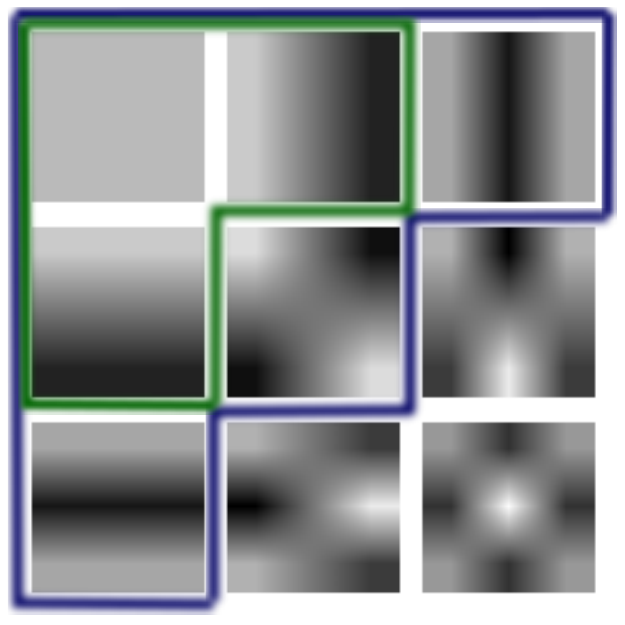}
\end{center}
\vspace{-1.5\baselineskip}
   \caption{DCT filter bank employed as feature extractor in the harmonic networks. Blue color marks filters used when $\lambda$=3, green color filters used to produce features at $\lambda$=2.}
\label{fig:filters}
\end{figure}

DCT transform is widely popular due to its energy compaction properties. The JPEG compression is motivated by the fact that in human vision the low frequency information is often prioritized over that contained in the high frequencies. Hence, JPEG encoding uses stronger quantization on higher frequency AC coefficients. We employ similar idea in the proposed harmonic network architecture. Specifically, we limit the visual spectrum of harmonic blocks to only several most valuable low frequencies, which results in a reduction of number of parameters and operations required at each block. The coefficients are  ordered by their importance for the visual system in triangular patterns starting in the zero frequency at the top-left corner, see Fig.~\ref{fig:filters}. We limit the spectrum of considered frequencies by hyperparameter $\lambda$ representing the number of levels of coefficients allowed along the main diagonal direction starting from zero frequency: DC only for $\lambda=1$, 3 coefficients (green) for $\lambda=2$, and 6 coefficients (purple) for $\lambda=3$. Eq.~\ref{eq:feature} is thus updated to:
\begin{equation} \label{eq:limited_feature}
\begin{aligned}
  & h^l = \sum_{n=0}^{N-1}{\sum_{u=0}^{\lambda-1}{\sum_{v=0}^{\lambda-1}{w^l_{n,u,v}\psi_{u,v}** h^{l-1}_n}}} \\
  & \text{subject to}\ u + v < \lambda; \lambda \leq K.
\end{aligned}
\end{equation}
Fig.~\ref{fig:filters} illustrates filters used at various levels assuming 3$\times$3 receptive field.

\subsection{Computational Requirements} \label{sec:method.requirements}

Harmonic blocks are designed to learn the same number of parameters as their convolutional counterparts. Requirements for the DCT transform scale linearly with the number of input channels and result in a modest increase to the theoretical number of operations. Standard convolutional layer used in many popular architectures that has $N$ input and $M$ output channels with a kernel size $K\times K$ learns $NMK^2$ parameters and performs $NMK^2AB$ operations if the filter is applied $A$ and $B$ times in particular directions. Harmonic block with $K^2$ transformation filters of size $K\times K$ upsamples representation to $NK^2$ features and then learns one weight for each upsampled-output feature pair hence $NK^2M$ weights. Transform of an $A\times B$ feature set costs $NK^2K^2AB$ on top of weighted combination $NK^2MAB$ that matches number of multiply-add operations of $K \times K$ convolution. The total number of operations is thus $NK^2AB\left(M+K^2\right)$. The theoretical number of multiply-add operations over the standard convolutional layer increases by a factor of ${K^2}/{M}$. If we assume truncated spectrum (use of $\lambda$) given by $P=\lambda(\lambda+1)/2$ filters, proportion of operations becomes $P/K^2+P/M$.

While keeping the number of parameters intact, a harmonic block requires additional memory during training and inference to store transformed feature representation. In our experiments with WRN models (Sec.\ref{sec:experiments.cifar}), the harmonic network trained with full DCT spectrum requires almost 3 times more memory than the baseline and may not fit onto a single GPU. This memory requirement can be eased by limiting the employed DCT spectrum.

Despite the comparable theoretical computational requirements, the run time of harmonic networks is larger compared to the baseline models, see Table~\ref{tab:requirements} below. This effect can be explained by the design of harmonic blocks that replaces a single convolutional layer by a block of 2 sequential convolutions (with harmonic filter and 1$\times$1 convolution). The model also handles larger feature space manipulating with more data. Finally, current hardware GPU accelerators are optimized to perform dense spatial convolutions and implementation of separable convolutions does not reach the full hardware capabilities. Harmonic network have the potential of running faster on dedicated hardware since they require less operations.

\section{Experiments} \label{sec:experiments}

\subsection{Small NORB dataset} \label{sec:experiments.norb}

The small NORB dataset~\cite{lecun04} is a synthetic set of binocular images of toys captured under different pose conditions: at 18 different angles, 9 elevations and 6 lighting conditions induced by combining different light sources. The set comprises 5 categories of toys, 5 instances of each category is used for training and 5 for testing, forming in total 24300 pairs of 96x96 grayscale images in each split. 

\vskip .1cm
\noindent{\bf Baseline.} The baseline CNN network is selected after an extensive hyper-parameter search. The network consists of 2 convolution layers with 32 5$\times$5 and 64 3$\times$3 filter banks respectively, a fully connected layer with 1024 neurons and a layer with softmax classifier. Filters are applied with stride 2 and features are subsampled by overlapping max-pooling. All hidden layer responses are batch normalized and rectified by ReLU. Moreover, we also use a slightly deeper network with an additional convolutional layer preceding the first pooling. Details of the architectures can be found in Table~\ref{tab:norb_nn}. The baseline CNNs are trained with stochastic gradient descent for 200 epochs with momentum 0.9. The initial learning rate 0.01 is decreased by factor 10 every 50 epochs. The network is trained with batches of 64 image pairs and each pair is padded with zeros 5 pixels on each side and a random crop of 96$\times$96 pixels is fed to the network. The optimization is regularized by dropout (p=0.5) on the fully connected layer and weight decay of 0.0005. 

\begin{table}
\begin{center}
\tabcolsep = .9mm
\begin{tabular}{c l l l}
\hline
Res. & CNN2 baseline & CNN3 baseline & Harmonic net\\
\hline
96x96 & conv 32,5x5/2 & conv 32,5x5/2 & harm 32,4x4/4 \\
48x48 & pool 3x3/2 & conv 64,3x3/2 & \\
24x24 & conv 64,3x3/2 & pool 2x2/2 & harm 64,3x3/2 \\
12x12 & pool 3x3/2 & conv 128,3x3/2 & pool 3x3/2 \\
6x6 & & pool 2x2/2 & harm 128,3x3/2\textsuperscript{*} \\
3x3 & fc 1024 & fc 1024 & fc 1024\textsuperscript{**} \\
1x1 & dropout 0.5 & dropout 0.5 & dropout 0.5 \\
1x1 & fc 5 & fc 5 & fc 5 \\
\hline
\end{tabular}
\end{center}
\vspace{-1\baselineskip}
\caption{Design of models used in small NORB experiments. Each row shows operation performed at given spatial resolution. For convolution and harmonic block we use notation \{conv, harm\}~M,K$\times$K/S with M output features, kernel size K and stride S. Similar notation is used for pooling: pool~K$\times$K/S and fully connected layer: fc~M. \textsuperscript{*}This layer is used only in the proposed network with 3 harmonic blocks. \textsuperscript{**}This fc layer can be replaced by harm 128,3x3/3.}
\label{tab:norb_nn}
\end{table}

Several versions of harmonic networks are considered, by substituting the first, first two or all three (in case of deeper network) convolution layers by harmonic blocks. Furthermore, the first fully-connected layer can be transformed to a harmonic block taking global DCT transform of the activations. The first harmonic block uses 4$\times4$ DCT filters applied without overlap, the further blocks mimic their convolutional counterparts with $3\times3$ kernels and stride~2.
Standard max or average pooling operator is applied between blocks. Using larger input receptive field realized by DCT transform did not provide any notable advantage.

\begin{figure}[t]
\begin{center}
   \includegraphics[width=\linewidth]{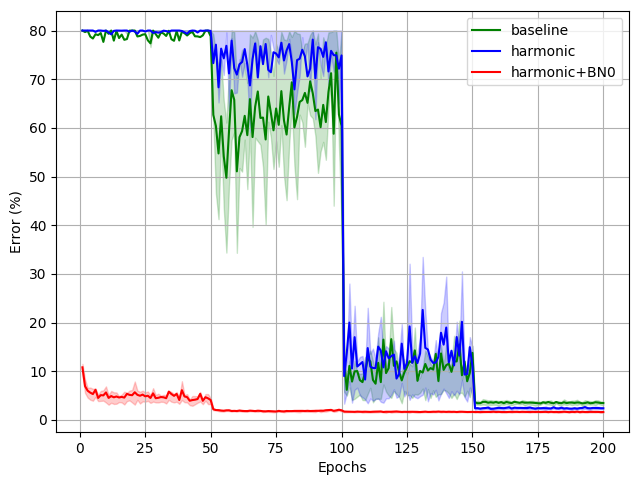}
\end{center}
\vspace{-1.5\baselineskip}
\caption{Mean classification error on small NORB test set. Weak generalization of CNN (green) and harmonic network (blue) is observed during the early stages of training. Filled areas (best seen in color) show 50\% empirical confidence intervals from 20 runs. Batch normalization of DCT spectrum (first block) significantly speeds up convergence of harmonic network (red).}
\label{fig:norb_convergence}
\end{figure}

\vskip .1cm
\noindent{\bf Results.} The baseline CNN architecture shows poor generalization performance in early stages of training. The performance on a test set stabilizes only after the third decrease of the learning rate, see Fig.~\ref{fig:norb_convergence}. Baseline CNN achieved mean error 3.48\%$\pm$0.50 from 20 trials, while CNN utilizing harmonic blocks without explicit normalization of harmonic responses exhibits similar behavior resulting in slightly better final solution 2.40\%$\pm$0.39. Normalizing DCT responses of input channels at the first block prevents harmonic network from focusing too much on pixel intensity, significantly speeds up generalization, improves performance and stability.

We observe the average pooling to work well in combination with harmonic blocks. Error of 1.14\%$\pm$0.20 is achieved by a hybrid model with one harmonic block followed by 2 standard convolutional layers using overlapping average pooling between them. A variant with 3 harmonic blocks and the same configuration performs comparably with 1.15\%$\pm$0.22 error (more results in Table~\ref{tab:norb_setting}). The best result was obtained by the model with 3 harmonic blocks replacing the convolutional layers and the fully-connected layer transformed into harmonic block, misclassifying only 1.10\%$\pm$0.16 of test samples. Turning the fully-connected layer used on spatial activations into the harmonic block provides translation invariance to the model.

\begin{table}
\begin{center}
\tabcolsep = 1mm
\begin{tabular}{ c c c c c }
\hline
Harm lay. & \#Conv lay. & Pooling type & Error (\%) \\
\hline
  & 2 & overlap max & 3.48$\pm$0.50 \\
  1 & 2 & max (no BN) & 2.40$\pm$0.39 \\
  1 & 2 & max & 1.63$\pm$0.19 \\
  1 & 2 & avg & 1.67$\pm$0.25 \\
  1,2 & 2 & max & 1.60$\pm$0.18\\
  1,2 & 2 & avg & 1.56$\pm$0.18 \\
  & 3 & max & 3.43$\pm$0.31 \\
  & 3 & overlap max & 3.89$\pm$0.65 \\
  1 & 3 & overlap avg (no BN) & 2.56$\pm$0.39 \\
  1 & 3 & overlap max & 1.16$\pm$0.15 \\
  1 & 3 & overlap avg & 1.14$\pm$0.20 \\
  1,2 & 3 & overlap max & 1.21$\pm$0.17 \\
  1,2 & 3 & overlap avg & 1.15$\pm$0.17 \\
  1,2,3 & 3 & overlap max & 1.18$\pm$0.16 \\
  1,2,3 & 3 & overlap avg & 1.15$\pm$0.22 \\
  1,2,3,4\textsuperscript{*} & 3 & overlap avg & \textbf{1.10$\pm$0.16} \\
\hline
\end{tabular}
\caption{Introduction of harmonic blocks into the baseline architectures and the respective errors on small NORB dataset. Networks with 2 and 3 conv/harm blocks have 2.39M and 1.28M parameters, respectively. The reduction is due to smaller fc layer. \textsuperscript{*}The fourth harm block replaces the fc layer, see Table~\ref{tab:norb_nn}.}
\label{tab:norb_setting}
\end{center}
\end{table}

Table~\ref{tab:norb_state_of_art} shows that these results surpass all previously reported error rates for this dataset to the best of our knowledge. The capsule network~\cite{Hinton18} claims 1.4\% error rate however estimated under different evaluation protocol, where testing images are resized to 48$\times$48 and prediction for multiple crops of size 32$\times$32 is averaged. The best reported result for a CNN~\cite{Ciresan11} 2.53\%$\pm$0.40 uses wider CNN architecture and four additional input channels derived from original input via contrast-extractive filters.

\begin{table}
\begin{center}
\begin{tabular}{ l c c c c }
\hline
Method & Num. Param. & Error (\%) \\
\hline
CNN with input filter~\cite{Ciresan11} & 2.7M & 2.53 $\pm$ 0.40 \\
Nonlinear SLPP~\cite{Rehn14} & & 1.5 \\
CapsNet~\cite{Hinton18} multi-crop & 0.31M & 1.4 \\
Harmonic net & 1.28M & \textbf{1.10 $\pm$ 0.16} \\
\hline
\end{tabular}
\end{center}
\vspace{-\baselineskip}
\caption{Comparison with the state-of-the-art on small NORB dataset, showing the proposed method outperforms other reported results.} 
\label{tab:norb_state_of_art}
\end{table}

The best performing model with three harmonic blocks and a global DCT transform of the final spatial feature is selected for designing a compact version of the model with a few modifications. The fully connected layer is reduced to only 32 neurons and the dropout is omitted. The modified network reaches 1.17\%$\pm$0.20 error and has 131k parameters. The harmonic blocks that extract only frequencies up to the level $\lambda=K$, equal to filter size, omit one third of weights for combining responses of 3x3 DCT filters. The network reaches 1.34\%$\pm$0.21 and needs to store less than 88k parameters. Going even further to learn only weights for frequencies where $\lambda=K-1$ reduces weights count to 3 per 3x3 filter, and totals in less than 45k parameters. This model uses non-overlapping pooling to preserve more of limited high-frequency information. Error of 1.64\%$\pm$0.22 is achieved on the test set, in contrast with small capsule network~\cite{Hinton18} with 68k parameters scoring 2.2\%.

Spectral representation of input data has a few interesting properties. Average feature intensity is captured into DC coefficient, while other coefficients (AC coefficients) capture signal structure. DCT representation has been successfully used~\cite{Er05} to build illumination invariant representation. This give us strong motivation to test illumination invariance properties of harmonic networks. The dataset is split into 3 parts based on lighting conditions during image capturing: the bright images (conditions 3,5) dark images (conditions 2,4) and images under standard lighting conditions (0,1). The models are trained only on data from one split and tested on  images with unseen lighting conditions. The baseline CNN performs poorly in terms of generalization. Using random brightness and contrast (only for dark images) augmentation improves the generalization and sometimes even validation error (on seen lighting conditions), most likely due to the limited amount of training data. Harmonic network as described in the previous paragraphs showed surprisingly poor performance compared to the baseline, even when using spectrum normalization. The model still pays too much attention to the DC component when making a decision. A good generalization performance is obtained when omitting the DC component from features on the input layer. Generalization errors of the best CNN and harmonic network architecture are reported in Table~\ref{tab:illum_inv} and Table~\ref{tab:illum_inv_uni} depicts generalization errors of models that have comparable validation errors, measured at different training stages. Even though the average intensity hinders the lighting invariance, it is still important for a competitive performance when training on the whole dataset. Brightness and contrast augmentation make a slight improvement to the generalization of harmonic networks, possibly by increasing the robustness of shape estimation from perturbed AC coefficients.

\begin{table}
\begin{center}
\tabcolsep = 0.5mm
\begin{tabular}{ l c c c c }
\hline
Augmentation & \multicolumn{2}{c}{No} & \multicolumn{2}{c}{Yes} \\
Lighting Cond. & CNN & Harmonic & CNN & Harmonic \\
\hline
Bright & 26.3$\pm$2.6 & 10.2$\pm$0.4 & 17.6$\pm$0.7 & 9.4$\pm$0.7\\
Standard & 30.2$\pm$1.8 & 18.0$\pm$1.9 & 22.48$\pm$1.1 & 15.5$\pm$1.1 \\
Dark & 31.2$\pm$1.8 & 18.9$\pm$1.2 & 20.1$\pm$1.3 & 14.5$\pm$1.4 \\
\hline
\end{tabular}
\end{center}
\vspace{-\baselineskip}
\caption{Generalization errors mean over 10 runs. Harmonic networks achieve lower error under every lighting condition both with and without brightness augmentation.} 
\label{tab:illum_inv}
\end{table}

\begin{table}
\begin{center}
\tabcolsep = 1mm
\begin{tabular}{ l c c c c }
\hline
Set & \multicolumn{2}{c}{Validation} & \multicolumn{2}{c}{Test} \\
Lighting Cond. & CNN & Harmonic & CNN & Harmonic \\
\hline
Bright & 7.80 & 7.80 & 17.82 & 16.40 \\
Standard & 7.77 & 8.23 & 31.05 & 26.86 \\
Dark & 8.79 & 9.38 & 20.21 & 18.79 \\
\hline
\end{tabular}
\end{center}
\vspace{-1\baselineskip}
\caption{Comparison of generalization to lighting conditions for models with similar validation errors (seen conditions). Harmonic networks still exhibit lower test error.} 
\label{tab:illum_inv_uni}
\end{table}

\begin{figure}[t]
\begin{center}
   \includegraphics[width=\linewidth]{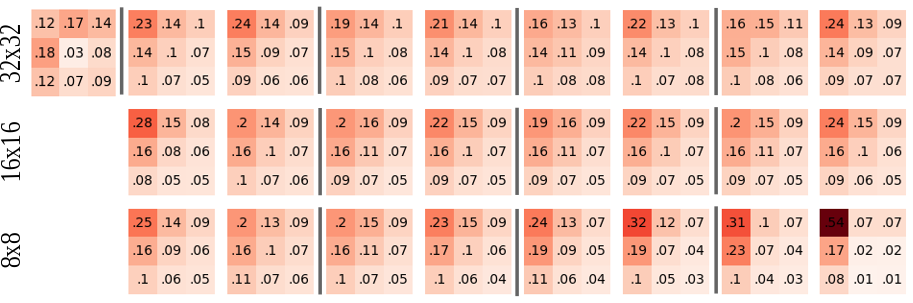}
\end{center}
\vspace{-1\baselineskip}
\caption{Distribution of weights assigned to DCT filters in the first harmonic block (left-most) and the remaining blocks in the fully harmonic WRN-28-10 trained on CIFAR10. Vertical lines separate the residual blocks.}
\label{fig:freq_heatmap}
\end{figure}

\begin{table}[!b]
\begin{center}
\tabcolsep = .75mm
\begin{tabular}{ l c c c c c c c c }
\hline
Method & dropout & CIFAR10 & CIFAR100 \\
\hline
WRN-28-10 & & 3.83 & 19.13 \\
WRN-28-10 & \checkmark & 3.91 & 18.75 \\
harm0 WRN-28-10 & & 4.1 & 19.17 \\
harm0+BN WRN-28-10 & & 3.9 & 18.8 \\
harm0+BN WRN-28-10 & \checkmark & 3.64 & 18.57 \\
fully harm WRN-28-10 & \checkmark & 3.86 & 18.57 \\
\hline
\end{tabular}
\end{center}
\vspace{-1\baselineskip}
\caption{Results achieved by modified harmonic wide residual network on CIFAR10 and CIFAR100. Errors shown are medians of 5 runs.} 
\label{tab:cifar_spec}
\end{table}

\begin{table*}
\begin{center}
\tabcolsep = 1.5mm
\begin{tabular}{ l c c c c c c c c }
\hline
Method & params & size & GPU mem. & TFlops & GPU time & CPU time & CIFAR10 & CIFAR100 \\
\hline
WRN-28-10 & 36.5M & 279MB & 5.8GB & 5.24 & 2.41 & 96.55 & 3.91 & 18.75 \\
WRN-28-8 & 23.4M & 179MB & 4.7GB & 3.36 & 1.68 & 63.84 & 4.01 & 19.38 \\
WRN-28-6 & 13.1M & 101MB & 3.4GB & 1.89 & 1.18 & 37.34 & 4.09 & 20.17 \\
harm0+BN WRN-28-10 & 36.5M & 279MB & 5.8GB & 5.24 & 2.41 & 95.17 & 3.64 & 18.57 \\
fully harm\;  WRN-28-10 & 36.5M & 279MB & 8.2+7.9GB & 5.4 & 5.18\textsuperscript{*} & 120.72 & 3.86 & 18.57 \\
fully harm\;  WRN-28-10 $\lambda$=3 & 24.4M & 187MB & 11.8GB & 3.62 & 5.73 & 88.41 & 3.84 & 18.58 \\
fully harm\;  WRN-28-10 $\lambda$=2 & 12.3M & 95MB & 8.1GB & 1.82 & 3.2 & 51.17 & 4.25 & 19.97 \\
\hline
\end{tabular}
\end{center}
\vspace{-1\baselineskip}
\caption{Computational requirements for harmonic WRNs and baselines of similar size. Fully harm WRN-28-10 $\lambda$=3 performs notably better than the baseline WRN-28-10 that has more parameters. CPU inference and GPU training time is reported per image in ms, estimated for the whole batch divided by the batch size 128, run on Intel Core i7-7700K CPU @ 4.20GHz and NVIDIA Titan X (Pascal) GPU. \textsuperscript{*}Time is estimated running the model on 2 GPUs.} 
\label{tab:requirements}
\end{table*}

\subsection{CIFAR datasets} \label{sec:experiments.cifar}

The second set of experiments is performed on popular benchmarking datasets of small natural images CIFAR10 and CIFAR100. Images have 3 color channels and resolution 32$\times$32 pixels. Train/test split is 50k images for training and 10k for testing. Images have balanced labeling, 10 classes in CIFAR10 and 100 in CIFAR100.

\vskip .1cm
\noindent{\bf Baseline.} For experiments on CIFAR datasets we adopt the Wide Residual Network~\cite{Zagoruyko16} with 28 layers and width multiplier 10 (WRN-28-10) as a baseline. Both the model design and training procedure are kept unchanged from the original paper. Harmonic WRNs are constructed by replacing convolutional layers by harmonic blocks with the same receptive field, preserving batch normalization and ReLU activations in their original positions after every block.

\vskip .1cm
\noindent{\bf Results.} The wide residual network selected as a baseline for CIFAR10 experiments improves over standard ResNets by using much wider residual blocks instead of extended depth. In our work we first investigate whether the WRN results can be improved if trained on spectral information, i.e. when replacing only the first convolutional layer preceding the residual blocks in WRN by a harmonic block with the same receptive field (harm0 WRN). The network learns more useful features if the RGB spectrum is explicitly normalized, surpassing the classification error of the baseline network on both CIFAR10 and CIFAR100 datasets, see Table~\ref{tab:requirements}. We then construct a fully harmonic WRN by replacing all convolutional layers with harmonic blocks, excluding residual shortcut projections. \textit{Zagoruyko et al.}~\cite{Zagoruyko16} demonstrated how a dropout used inside a residual block placed between convolutional layers can provide extra regularization when trained on spatial data~\cite{Zagoruyko16}. This conforms with our findings when training on spectral representations, therefore we adopt dropout between harmonic blocks. The harmonic network outperforms the baseline WRN, providing the performance gains as reported in Table~\ref{tab:cifar_spec}.

\begin{figure}[t]
\begin{center}
   \includegraphics[width=\linewidth]{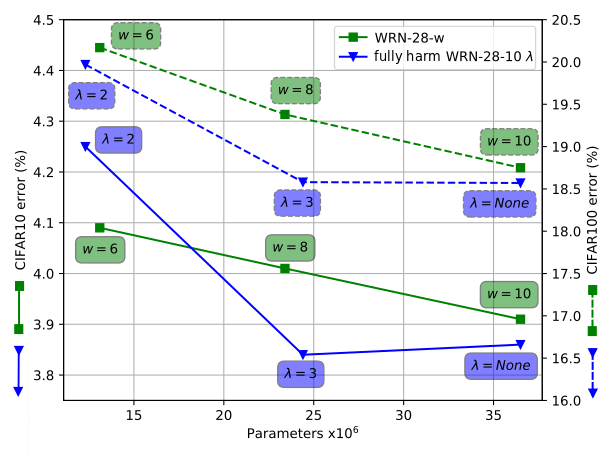}
\end{center}
\vspace{-1.5\baselineskip}
\caption{Graphs show a decrease of classification error as a function of model size on CIFAR10 (solid lines) and CIFAR100 (dahsed lines). Parameters of harmonic networks are modulated by spectrum size $\lambda$, the WRN baselines are controlled by the width multiplier {\it w}.}
\label{fig:compression_graph}
\end{figure}

Analysis of fully harmonic WRN weights learned with 3$\times$3 spectrum revealed that the deeper network layers tend to favour low-frequency information over high frequencies when learning representations. Relative importance of weights corresponding to different frequencies shown in Fig.~\ref{fig:freq_heatmap} motivates truncation of high-frequency coefficients for compression purposes. While preserving the input image spectrum intact, we train the harmonic networks on limited spectrum of hidden features for $\lambda$=2 and $\lambda$=3 using 3 and 6 DCT filters for each feature, respectively. Fully harmonic WRN with $\lambda$=3 has approximately 33\% less parameters with respect to the baseline, while fully harm WRN with $\lambda$=2 is roughly 3 times smaller. To assess the loss of accuracy associated with parameter reduction we train baselines with reduced widths having comparable numbers of parameters: WRN-28-8 and WRN-28-6, see Fig.~\ref{fig:compression_graph}. Fully harmonic WRN-28-10 with $\lambda$=3 has comparable error to the network using the full spectrum and outperforms the larger baseline WRN-28-10, showing almost no loss in discriminatory information. On the other hand fully harmonic WRN-28-10 with $\lambda$=2 is better on CIFAR100 and slightly worse on CIFAR10 compared to the similarly sized WRN-28-6. The performance degradation indicates that some of the truncated coefficients carry important discriminatory information. Detailed comparison including computational requirements is reported in Table~\ref{tab:requirements}. Notable increase in training time and memory requirements of harmonic networks is due to reasons described in Section~\ref{sec:method.requirements}. Nevertheless, the decrease in the number of operations for compressed harmonic networks can be seen in CPU inference time since CPU is not biased by implementation of parallel operations.

\section{Conclusion}

We have presented a novel approach to incorporate spectral information from DCT into CNNs by designing a harmonic architectural block that extracts per-pixel spectral information from input and intermediate features and learns weights to combine this information to form new representations. We empirically validate the use of harmonic blocks by integrating them into the well-established CNN architectures to validate the improvement associated with the proposed CNN block in terms of both classification accuracy and parametric complexity. We also demonstrate illumination invariance properties of networks with harmonic blocks and reduction in model size without degradation in performance by truncation of parameters corresponding to high frequencies of feature visual spectrum. Since by design harmonic networks capture periodicity in spatial domain, these networks are expected to be particularly impactful for tasks such as texture classification.

\section*{Acknowledgement}
 This research was supported by the ADAPT Centre for Digital Content Technology funded under the SFI Research Centres Programme (Grant 13/RC/2106) and co-funded under the European Regional Development Fund. The second author was also supported by the European Union`’s Horizon 2020 research and innovation programme under the Marie Sklodowska-Curie grant agreement No.713567. We gratefully acknowledge the support of NVIDIA Corporation with the donation of the Titan Xp GPUs used for this research.

{\small
\bibliographystyle{ieee}
\bibliography{bib}

\begin{thebibliography}{10}\itemsep=-1pt

\bibitem{Amerini17}
I.~Amerini, T.~Uricchio, L.~Ballan, and R.~Caldelli.
\newblock Localization of {JPEG} double compression through multi-domain
  convolutional neural networks.
\newblock In {\em 2017 IEEE Conference on Computer Vision and Pattern
  Recognition Workshops (CVPRW)}, pages 1865--1871, July 2017.

\bibitem{Barni17}
M.~Barni, L.~Bondi, N.~Bonettini, P.~Bestagini, A.~Costanzo, M.~Maggini,
  B.~Tondi, and S.~Tubaro.
\newblock Aligned and non-aligned double {JPEG} detection using convolutional
  neural networks.
\newblock {\em J. Vis. Comun. Image Represent.}, 49(C):153--163, Nov. 2017.

\bibitem{Bruna13}
J.~Bruna and S.~Mallat.
\newblock Invariant scattering convolution networks.
\newblock {\em IEEE transactions on pattern analysis and machine intelligence},
  35(8):1872--1886, 2013.

\bibitem{Chen15}
W.~Chen, J.~Wilson, S.~Tyree, K.~Weinberger, and Y.~Chen.
\newblock Compressing neural networks with the hashing trick.
\newblock In {\em International Conference on Machine Learning}, pages
  2285--2294, 2015.

\bibitem{Chen16}
W.~Chen, J.~Wilson, S.~Tyree, K.~Q. Weinberger, and Y.~Chen.
\newblock Compressing convolutional neural networks in the frequency domain.
\newblock In {\em Proceedings of the 22Nd ACM SIGKDD International Conference
  on Knowledge Discovery and Data Mining}, KDD '16, pages 1475--1484, New York,
  NY, USA, 2016. ACM.

\bibitem{Ciresan11}
D.~C. Ciresan, U.~Meier, J.~Masci, L.~Maria~Gambardella, and J.~Schmidhuber.
\newblock Flexible, high performance convolutional neural networks for image
  classification.
\newblock In {\em IJCAI Proceedings-International Joint Conference on
  Artificial Intelligence}, volume~22, page 1237. Barcelona, Spain, 2011.

\bibitem{Denil13}
M.~Denil, B.~Shakibi, L.~Dinh, M.~A. Ranzato, and N.~de~Freitas.
\newblock Predicting parameters in deep learning.
\newblock In C.~J.~C. Burges, L.~Bottou, M.~Welling, Z.~Ghahramani, and K.~Q.
  Weinberger, editors, {\em Advances in Neural Information Processing Systems
  26}, pages 2148--2156. Curran Associates, Inc., 2013.

\bibitem{Er05}
M.~J. Er, W.~Chen, and S.~Wu.
\newblock High-speed face recognition based on discrete cosine transform and
  rbf neural networks.
\newblock {\em Trans. Neur. Netw.}, 16(3):679--691, May 2005.

\bibitem{Fu16}
D.~Fu and G.~Guimaraes.
\newblock Using compression to speed up image classification in artificial
  neural networks.
\newblock 2016.

\bibitem{Fujieda17}
S.~Fujieda, K.~Takayama, and T.~Hachisuka.
\newblock Wavelet convolutional neural networks for texture classification.
\newblock {\em arXiv preprint arXiv:1707.07394}, 2017.

\bibitem{Ghosh16}
A.~Ghosh and R.~Chellappa.
\newblock Deep feature extraction in the dct domain.
\newblock In {\em 2016 23rd International Conference on Pattern Recognition
  (ICPR)}, pages 3536--3541, Dec 2016.

\bibitem{Gueguen18}
L.~Gueguen, A.~Sergeev, R.~Liu, and J.~Yosinski.
\newblock Faster neural networks straight from {JPEG}.
\newblock In {\em International Conference on Learning Representations
  Workshop}, 2018.

\bibitem{Han16}
S.~Han, H.~Mao, and W.~J. Dally.
\newblock Deep compression: Compressing deep neural networks with pruning,
  trained quantization and huffman coding.
\newblock {\em International Conference on Learning Representations (ICLR)},
  2016.

\bibitem{Hinton18}
G.~E. Hinton, S.~Sabour, and N.~Frosst.
\newblock Matrix capsules with {EM} routing.
\newblock In {\em International Conference on Learning Representations}, 2018.

\bibitem{Jacobsen16}
J.-H. Jacobsen, J.~van Gemert, Z.~Lou, and A.~W. Smeulders.
\newblock Structured receptive fields in cnns.
\newblock In {\em Proceedings of the IEEE Conference on Computer Vision and
  Pattern Recognition}, pages 2610--2619, 2016.

\bibitem{Jaderberg14}
M.~Jaderberg, A.~Vedaldi, and A.~Zisserman.
\newblock Speeding up convolutional neural networks with low rank expansions.
\newblock In {\em Proceedings of the British Machine Vision Conference}. BMVA
  Press, 2014.

\bibitem{Kim15}
M.~Kim and L.~Rigazio.
\newblock Deep clustered convolutional kernels.
\newblock In {\em Feature Extraction: Modern Questions and Challenges}, pages
  160--172, 2015.

\bibitem{lecun04}
Y.~LeCun, F.~J. Huang, and L.~Bottou.
\newblock Learning methods for generic object recognition with invariance to
  pose and lighting.
\newblock In {\em Computer Vision and Pattern Recognition, 2004. CVPR 2004.
  Proceedings of the 2004 IEEE Computer Society Conference on}, volume~2, pages
  II--104. IEEE, 2004.

\bibitem{Li17}
B.~Li, H.~Luo, H.~Zhang, S.~Tan, and Z.~Ji.
\newblock A multi-branch convolutional neural network for detecting double
  {JPEG} compression.
\newblock {\em CoRR}, abs/1710.05477, 2017.

\bibitem{Lu18}
H.~Lu, H.~Wang, Q.~Zhang, D.~Won, and S.~W. Yoon.
\newblock A dual-tree complex wavelet transform based convolutional neural
  network for human thyroid medical image segmentation.
\newblock In {\em 2018 IEEE International Conference on Healthcare Informatics
  (ICHI)}, pages 191--198, June 2018.

\bibitem{Mathieu14}
M.~Mathieu, M.~Henaff, and Y.~Lecun.
\newblock Fast training of convolutional networks through ffts.
\newblock In {\em International Conference on Learning Representations
  (ICLR2014), CBLS, April 2014}, 2014.

\bibitem{Oyallon17}
E.~Oyallon, E.~Belilovsky, and S.~Zagoruyko.
\newblock Scaling the scattering transform: Deep hybrid networks.
\newblock In {\em The IEEE International Conference on Computer Vision (ICCV)},
  Oct 2017.

\bibitem{Oyallon18}
E.~Oyallon, E.~Belilovsky, S.~Zagoruyko, and M.~Valko.
\newblock Compressing the input for cnns with the first-order scattering
  transform.
\newblock In {\em ECCV 2018-European Conference on Computer Vision}, 2018.

\bibitem{Oyallon15}
E.~Oyallon and S.~Mallat.
\newblock Deep roto-translation scattering for object classification.
\newblock In {\em The IEEE Conference on Computer Vision and Pattern
  Recognition (CVPR)}, June 2015.

\bibitem{Perronnin15}
F.~Perronnin and D.~Larlus.
\newblock Fisher vectors meet neural networks: A hybrid classification
  architecture.
\newblock In {\em The IEEE Conference on Computer Vision and Pattern
  Recognition (CVPR)}, June 2015.

\bibitem{Rehn14}
E.~M. Rehn and H.~Sprekeler.
\newblock Nonlinear supervised locality preserving projections for visual
  pattern discrimination.
\newblock In {\em 2014 22nd International Conference on Pattern Recognition},
  pages 1568--1573, Aug 2014.

\bibitem{Rippel15}
O.~Rippel, J.~Snoek, and R.~P. Adams.
\newblock Spectral representations for convolutional neural networks.
\newblock In {\em Advances in neural information processing systems}, pages
  2449--2457, 2015.

\bibitem{Said16}
S.~Said, O.~Jemai, S.~Hassairi, R.~Ejbali, M.~Zaied, and C.~B. Amar.
\newblock Deep wavelet network for image classification.
\newblock In {\em 2016 IEEE International Conference on Systems, Man, and
  Cybernetics}, pages 922--927, Oct 2016.

\bibitem{Silva18}
D.~D. N.~D. Silva, S.~Fernando, I.~T.~S. Piyatilake, and A.~V.~S. Karunarathne.
\newblock Wavelet based edge feature enhancement for convolutional neural
  networks, 2018.

\bibitem{Simonyan13}
K.~Simonyan, A.~Vedaldi, and A.~Zisserman.
\newblock Deep fisher networks for large-scale image classification.
\newblock In C.~J.~C. Burges, L.~Bottou, M.~Welling, Z.~Ghahramani, and K.~Q.
  Weinberger, editors, {\em Advances in Neural Information Processing Systems
  26}, pages 163--171. Curran Associates, Inc., 2013.

\bibitem{Singh17}
A.~Singh and N.~Kingsbury.
\newblock Efficient convolutional network learning using parametric log based
  dual-tree wavelet scatternet.
\newblock In {\em Computer Vision Workshop (ICCVW), 2017 IEEE International
  Conference on}, pages 1140--1147. IEEE, 2017.

\bibitem{Torfason18}
R.~Torfason, F.~Mentzer, E.~Agustsson, M.~Tschannen, R.~Timofte, and L.~V.
  Gool.
\newblock Towards image understanding from deep compression without decoding.
\newblock In {\em International Conference on Learning Representations}, 2018.

\bibitem{Ulicny17}
M.~Ulicny and R.~Dahyot.
\newblock On using {CNN} with {DCT} based image data.
\newblock In {\em Irish Machine Vision and Image Processing Conference}, 2017.

\bibitem{Wang16}
Q.~Wang and R.~Zhang.
\newblock Double {JPEG} compression forensics based on a convolutional neural
  network.
\newblock {\em EURASIP Journal on Information Security}, 2016(1):23, Oct 2016.

\bibitem{Wang16b}
Y.~Wang, C.~Xu, S.~You, D.~Tao, and C.~Xu.
\newblock Cnnpack: Packing convolutional neural networks in the frequency
  domain.
\newblock In D.~D. Lee, M.~Sugiyama, U.~V. Luxburg, I.~Guyon, and R.~Garnett,
  editors, {\em Advances in Neural Information Processing Systems 29}, pages
  253--261. Curran Associates, Inc., 2016.

\bibitem{Williams16}
T.~Williams and R.~Li.
\newblock Advanced image classification using wavelets and convolutional neural
  networks.
\newblock In {\em 2016 15th IEEE International Conference on Machine Learning
  and Applications (ICMLA)}, pages 233--239, Dec 2016.

\bibitem{Williams18}
T.~Williams and R.~Li.
\newblock Wavelet pooling for convolutional neural networks.
\newblock In {\em International Conference on Learning Representations}, 2018.

\bibitem{Worrall17}
D.~E. Worrall, S.~J. Garbin, D.~Turmukhambetov, and G.~J. Brostow.
\newblock Harmonic networks: Deep translation and rotation equivariance.
\newblock In {\em Computer Vision and Pattern Recognition (CVPR), 2017 IEEE
  Conference on}, pages 7168--7177. IEEE, 2017.

\bibitem{Yaroslavsky14}
L.~P. Yaroslavsky.
\newblock Fast transforms in image processing: compression, restoration, and
  resampling.
\newblock {\em Advances in Electrical Engineering}, 2014, 2014.

\bibitem{Zagoruyko16}
S.~Zagoruyko and N.~Komodakis.
\newblock Wide residual networks.
\newblock In E.~R.~H. Richard C.~Wilson and W.~A.~P. Smith, editors, {\em
  Proceedings of the British Machine Vision Conference (BMVC)}, pages
  87.1--87.12. BMVA Press, September 2016.

\bibitem{Zou14}
X.~Zou, X.~Xu, C.~Qing, and X.~Xing.
\newblock High speed deep networks based on discrete cosine transformation.
\newblock In {\em Image Processing (ICIP), 2014 IEEE International Conference
  on}, pages 5921--5925. IEEE, 2014.

\end{thebibliography}
}

\end{document}